\documentclass[10pt,twocolumn,letterpaper]{article}

\usepackage[pagenumbers]{cvpr} 
\usepackage{multirow}

\definecolor{cvprblue}{rgb}{0.21,0.49,0.74}
\usepackage[pagebackref,breaklinks,colorlinks,allcolors=cvprblue]{hyperref}
\usepackage[T1]{fontenc}
\usepackage[utf8]{inputenc}
\usepackage{tcolorbox}
\usepackage{xcolor}
\tcbuselibrary{listings}
\newtcblisting{promptbox}{
    colback=gray!5,
    colframe=gray!50,
    arc=3pt,
    boxrule=0.5pt,
    listing only,
    left=5pt,
    right=5pt,
    top=5pt,
    bottom=5pt,
    listing options={
        basicstyle=\small\ttfamily,
        breaklines=true,
        columns=flexible,
        keepspaces=true,
        escapeinside={(*}{*)}
    }
}

\def\confName{CVPR}
\def\confYear{2026}

\title{MSUE: Multi-Modal Soccer Understanding Expert}

\author{%
\parbox{\linewidth}{\centering
\textbf{Litao Li}\textsuperscript{1,\textdagger},
\textbf{Yibo Yu}\textsuperscript{2},
\textbf{Yufeng Hu}\textsuperscript{4},
\textbf{Zhuo Yang}\textsuperscript{3},
\textbf{Jiali Wen}\textsuperscript{1},
\textbf{Yixin Chen}\textsuperscript{1},
\textbf{Yixi Zhou}\textsuperscript{1}\\[2pt]
{
\textsuperscript{1}South China University of Technology,
\textsuperscript{2}Johns Hopkins University,
\textsuperscript{3}Peking University\\
\textsuperscript{4}University of Electronic Science and Technology of China\\
\textsuperscript{\textdagger}Project lead \quad {\tt\small seonyee@foxmail.com}
}
}
}

\begin{document}
\maketitle
\begin{abstract}
This paper presents our solution to the 2026 SoccerNet VQA Challenge. We first develop a cost-effective data synthesis pipeline driven by a Vision-Language Model (VLM), which systematically restructures raw domain data into diverse VQA samples, including concise answers and long-form responses. Second, we propose MSUE, a multi-expert question answering architecture that employs a Large Language Model (LLM) to dynamically dispatch questions to text, image, and video experts. These experts are instantiated as a strong text baseline Gemini3-Flash, a fine-tuned Qwen3-VL, and an external knowledge base, respectively, working collaboratively to enhance VQA performance. MSUE achieves an accuracy of \textbf{0.95} on the challenge benchmark, securing third place in the leaderboard.
\end{abstract}

\section{Introduction}
\label{sec:intro}
Visual Question Answering (VQA) in dynamic soccer scenarios is a challenging task that requires precise visual perception, temporal reasoning, and rich domain knowledge. Existing approaches often struggle to jointly handle heterogeneous modalities and domain-specific reasoning, leading to limited performance in complex real-world settings. In this work, we present our winning solution for the 2026 SoccerNet VQA Challenge. 

We first construct a cost effective data synthesis pipeline driven by a Vision Language Model, which systematically transforms raw domain data into diverse VQA samples with both concise and long form answers, significantly improving data coverage and diversity. Building upon this, we propose MSUE, a multi expert question answering framework that leverages a Large Language Model to dynamically route queries to specialized experts in text, image, and video domains, including Gemini3-Flash, a fine-tuned Qwen3-VL, and an external knowledge base. This collaborative design enables effective integration of multimodal perception and domain knowledge. Our approach achieves an accuracy of \textbf{0.95} on the challenge benchmark, surpassing the second place by over 5\%, establishing a new state of the art and demonstrating the effectiveness of our framework.

\section{Data Curation}
\label{sec:data_pipeline}

\subsection{Data Source}
Our training pipeline leverages data from SoccerBench~\cite{soccerbench}, SoccerNet-V3~\cite{soccernet,soccernettracking,Cioppa2022,soccernetv2}, SoccerChat~\cite{soccerchat}, SoccerReplay-1988~\cite{soccerreplay}, SoccerNet-Caption~\cite{sn-caption}, and SoccerWiki~\cite{soccerbench}. After manual cleaning and consistency rectification, we obtain 32,151 high-quality instruction-tuning samples. The overall workflow is summarized in Figure~\ref{basicins}.

\subsection{Data Pipeline}

\textbf{Caption and MCQ Generation.} The raw corpus contains both pure textual sources and visual data, but their supervision signals are not directly aligned in a unified training format. To construct dense semantic annotations, we first use DeepSeek-V3.2~\cite{deepseekv32} and Qwen3-VL-32B~\cite{qwen3} to derive detailed captions from these heterogeneous inputs, covering both textual knowledge and fine-grained visual content. Based on the resulting dense captions, we further employ Qwen3-VL-32B to generate multiple-choice questions (MCQs) together with candidate options and reference answers, thereby converting all samples into a standardized VQA MCQ format.

\textbf{Ensemble Consistency Verification.} Initial reformatted data contains ~20\% noise. We use three VLMs: Step3-VL-10B~\cite{step3vl}, Qwen3-VL-235B-A22B~\cite{qwen3}, and InternVL3.5-241B-A28B~\cite{internvl35} for cross-validation. A sample is kept only if at least two models agree on the answer. For ambiguous samples, 80\% are discarded to reduce label noise, while 20\% are retained to preserve diversity.

\subsection{Task-Guided Data Synthesis}
To increase instructional diversity, we perform SFT with LoRA~\cite{lora} on Qwen3-VL-32B to build an instruction engine. This engine converts single-query instances into multi-turn dialogues by projecting 14 predefined QA paradigms onto each visual input, generating roughly three diverse reasoning tasks per sample. The prompt design is illustrated in Figure~\ref{basicins}.

\section{Baseline}
\label{sec:baseline}
To determine the optimal baseline for subsequent training, we evaluate the performance of several state-of-the-art (SOTA) Vision-Language Models (VLMs) on the official test set. The comparative results are summarized in Table~\ref{baselines}.

\begin{table}[h]
\centering
\caption{Performance comparison of various SOTA VLMs on the official SoccerNet test set. The best results are highlighted in \textbf{bold}.}
\label{baselines}
\begin{tabular}{lcc}
\toprule
Model & Params & Accuracy (\%) \\
\midrule
GPT-4o~\cite{openai2024gpt4ocard} & - & 0.39 \\
Qwen2.5-VL-7B~\cite{qwen25} & 7B & 0.34 \\
Qwen2.5-VL-32B~\cite{qwen25} & 32B & 0.40 \\
InternVL3.5-38B~\cite{internvl35} & 38B & 0.41 \\
Step3-VL-10B~\cite{step3vl} & 10B & 0.37 \\
Qwen3-VL-8B~\cite{qwen3} & 8B & 0.38 \\ \hline
\textbf{Qwen3-VL-32B}~\cite{qwen3} & 32B & \textbf{0.45} \\
\bottomrule
\end{tabular}
\end{table}

Based on the results, we selected Qwen3-VL-32B, which exhibits superior zero-shot generalization capabilities, as our baseline.

\section{Method}
\label{sec:method}
\begin{figure*}[!ht]
\centering
\includegraphics[width=\textwidth]{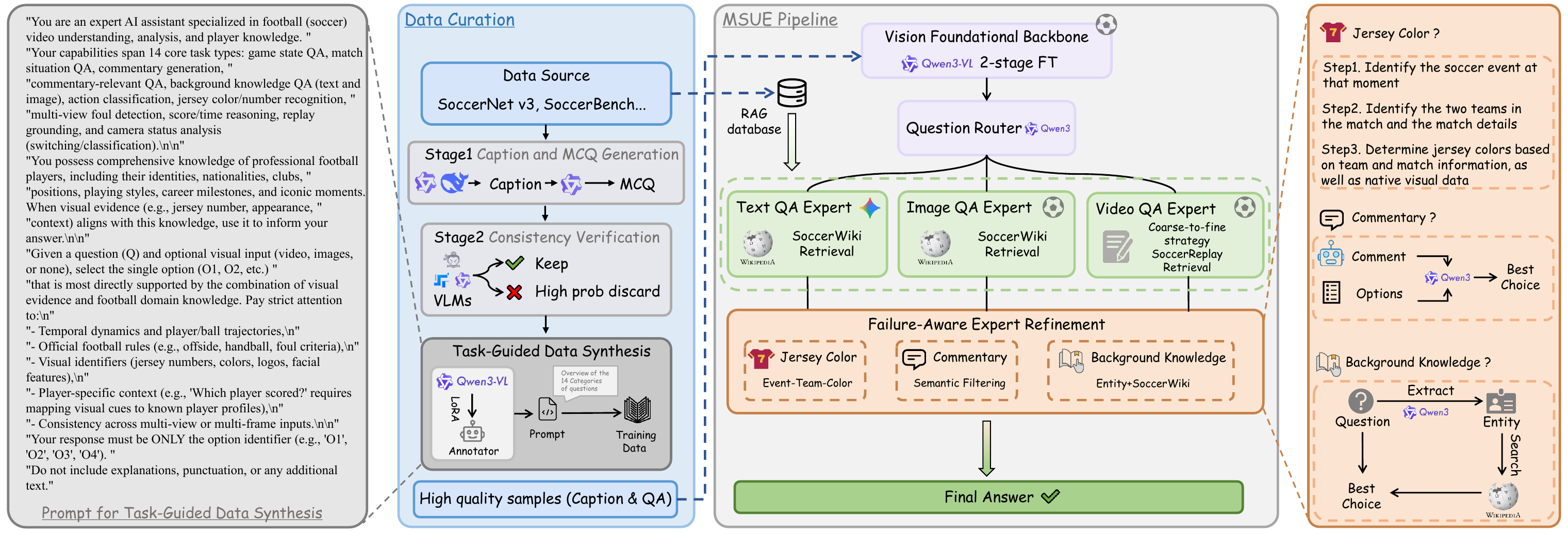}
\caption{Overview of our overall framework, including the data curation pipeline, the prompt design for task-guided data synthesis, the working pipeline of MSUE, and the failure-aware expert refinement strategy.}
\label{basicins}
\end{figure*}

Inspired by SoccerAgent~\cite{soccerbench}, We propose \textbf{MSUE} (Multi-modal Soccer Understanding Expert), a modular divide-and-conquer framework for football video, image, and text understanding. MSUE consists of a domain-adapted vision backbone, a lightweight question router, and specialized experts for text, image, and video reasoning. By decomposing heterogeneous VQA questions into expert-specific inference paths, the framework improves robustness and efficiency under diverse reasoning demands. Our method achieves 0.95 accuracy on the challenge benchmark. The overall framework is illustrated in Figure~\ref{basicins}.

\subsection{Vision Foundational Backbone}
The Vision Foundational Backbone (VFB) is built upon Qwen3-VL-32B-Instruct~\cite{qwen3} through a two-stage adaptation pipeline. We first perform full-parameter tuning on a curated football caption corpus to inject domain-specific visual semantics. We then conduct LoRA instruction tuning on augmented VQA data spanning open-ended and multiple-choice tasks. This progressive adaptation strengthens football reasoning while preserving general multimodal capability, yielding a strong unified visual expert.
We implement the two-stage fine-tuning pipeline using the ms-swift and execute all training and inference procedures on a compute node equipped with 8$\times$NVIDIA H100 GPUs. The specific configurations for our two-stage VFB training pipeline are detailed in Table~\ref{train-detail}.
\begin{table}[t]
\footnotesize
\centering
\caption{Training Hyper-parameters for the two-stage VFB training pipeline.}
\label{train-detail}
\begin{tabular}{@{}lcc@{}}
\toprule
\textbf{Hyper-param} & \textbf{Stage 1: Captioning} & \textbf{Stage 2: Q\&A} \\ \midrule
Type         & Full-parameter               & LoRA (Rank=16,Alpha=32)    \\
Learning Rate            & $1 \times 10^{-5}$           & $2 \times 10^{-4}$            \\
Epochs                   & 2                            & 4                             \\
Batch Size     & 8                            & 16                            \\
DeepSpeed     & Zero3-offload                            & Zero2                            \\
max-length     & 8192                           & 10240                            \\
Weight Decay             & 0.05                         & 0.1                           \\
Warmup-Ratio             & 0.05                         & 0.05                           \\ \bottomrule
\end{tabular}
\end{table}
\subsection{Question Routing and Expert Dispatch}
To address the heterogeneous nature of soccer VQA, we directly use Qwen3-4B~\cite{qwen3} with a routing prompt to classify each query into 14 predefined categories and then dispatch it to the corresponding expert. This routing stage separates questions requiring temporal reasoning, visual grounding, knowledge retrieval, or multi-step inference before answer generation, substantially reducing errors caused by inappropriate inference paths.

\subsection{Multi-Modal Q\&A Experts}
Following question routing, MSUE dispatches each query to one of three modality-specific experts. The \textbf{Text Q\&A Expert} uses Gemini-3-Flash~\cite{gemini} with SoccerWiki retrieval. Structured knowledge entries are encoded by all-MiniLM-L6-v2, and the retrieved evidence is appended to the prompt for grounded answer generation.

The \textbf{Image Q\&A Expert} uses the fine-tuned VFB with cross-modal retrieval over SoccerWiki images and metadata. The retrieved references provide external evidence for entity recognition and scene understanding.

The \textbf{Video Q\&A Expert} uses the fine-tuned VFB with a coarse-to-fine strategy. The model first produces dense descriptions of entities, actions, and scene context. These intermediate summaries are then used to retrieve matched evidence from SoccerReplay~\cite{soccerreplay}, which is incorporated through retrieval-augmented generation for final answer prediction.

\subsection{Failure-Aware Expert Refinement}
Beyond the core expert architecture, we introduce targeted refinements for representative failure-prone categories.

For \textit{Jersey Color Relevant QA}, we decompose reasoning into event recognition, responsible-player identification, and jersey color localization, reducing confusion in visually similar scenarios. For \textit{Commentary Generation}, we address hallucination and answer inconsistency by introducing semantic alignment between generated commentary and candidate options, filtering irrelevant choices before final prediction. For \textit{Background Knowledge Image QA}, we first identify the queried entity type, then retrieve matched SoccerWiki knowledge for prompt augmentation, improving disambiguation and reducing errors caused by incomplete parametric knowledge.

These refinements complement expert routing and improve robustness on cases involving ambiguity, hallucination, and external knowledge dependence.

\section{Experiments}
\subsection{Implementation Details}
All training and inference procedures were executed on a high-performance compute node equipped with 8$\times$ NVIDIA H100 GPUs. The specific configurations for our two-stage VFB training pipeline are detailed in Table~\ref{train-detail}.
\begin{table}[t]
\footnotesize
\centering
\caption{Training Hyper-parameters for the two-stage VFB training pipeline. All experiments were conducted on a server with 8$\times$ NVIDIA H100 GPUs.}
\label{train-detail}
\begin{tabular}{@{}lcc@{}}
\toprule
\textbf{Hyper-param} & \textbf{Stage 1: Captioning} & \textbf{Stage 2: Q\&A (LoRA)} \\ \midrule
Type         & Full-parameter               & LoRA    \\
Learning Rate            & $1 \times 10^{-5}$           & $2 \times 10^{-4}$            \\
Epochs                   & 2                            & 4                             \\
Batch Size     & 8                            & 16                            \\
LoRA Rank ($r$)          & -                            & 16                            \\
LoRA Alpha ($\alpha$)    & -                            & 32                            \\
Weight Decay             & 0.05                         & 0.1                           \\
Warmup-Ratio             & 0.05                         & 0.05                           \\ \bottomrule
\end{tabular}
\end{table}

\subsection{Comparison Experiments}
We evaluate M-SUE on the SoccerNet-VQA test set and compare it with several state-of-the-art MLLMs. As shown in Table~\ref{tab:main_results}, our method achieves a top accuracy of \textbf{0.91}, setting a new state-of-the-art performance. The results demonstrate that while large-scale general VLMs (e.g., InternVL3.5) possess strong foundational capabilities, they often struggle with specialized soccer rules and player identification in zero-shot scenarios. By leveraging our high-quality data synthesis pipeline, M-SUE successfully aligns professional soccer domain knowledge with visual perception, significantly outperforming both base models and larger-scale competitors.

\begin{table}[ht]
\centering
\small
\caption{Comparison with state-of-the-art methods on the SoccerNet-VQA test set. FT and ZS denote Fine-tuned and Zero-shot, respectively. Our M-SUE achieves the best performance among all models.}
\label{tab:main_results}
\begin{tabular}{@{}lccc@{}}
\toprule
Method & Type & Parameters & Accuracy (\%) \\ \midrule
Qwen2.5-VL-7B         & FT & 7B  & 0.54 \\
Qwen2.5-VL-32B        & FT & 32B & 0.63 \\
InternVL3.5-38B       & FT & 38B & 0.63 \\
Qwen3-VL-8B           & FT & 8B  & 0.6 \\
Qwen3-VL-32B  & FT & 32B & 0.78 \\ \midrule
Qwen3-VL-235B-A22B    & ZS & 235B & 0.65 \\
InternVL3.5-241B-A28B & ZS & 241B & 0.64 \\
Qwen3.5-397B-A17B     & ZS & 397B & 0.72 \\ \midrule
\textbf{M-SUE (Ours)} & \textbf{FT} & \textbf{32B} & \textbf{0.91} \\ \bottomrule
\end{tabular}
\end{table}

\subsection{Ablation on VFB Training Stages}
We evaluate the incremental gains of the VFB across different training phases. Table~\ref{tab:ablation_vfb} illustrates the performance trajectory from general perception to domain-specific reasoning.

\begin{table}[ht]
\centering
\small
\caption{Ablation study of VFB training stages on the SoccerNet-VQA validation set. Stage 1 aligns basic soccer visual features, while Stage 2 introduces large-scale synthetic instruction data.}
\label{tab:ablation_vfb}
\begin{tabular}{@{}lcc@{}}
\toprule
Training Stage & Data Composition & Accuracy (\%) \\ \midrule
Baseline & - & 0.45 \\
Stage 1    & Caption Pairs     & 0.73 \\
Stage 2 & Q\&A Pairs & \textbf{0.78} \\ \bottomrule
\end{tabular}
\end{table}

As shown, the Baseline model exhibits a substantial domain gap. Stage 1 (Caption Pairs) bridges this gap by establishing a mapping between visual actions and professional soccer vocabulary, yielding a \textbf{+28\% gain}. Stage 2 (Q\&A Pairs) further improves accuracy to \textbf{0.78}, indicating that instruction-tuning on diverse question formats is essential for evolving from passive description to active temporal reasoning.

\subsection{Effectiveness of Expert Design}
We conduct a granular ablation study to evaluate the individual contributions of the Text, Image, and Video Experts. As shown in Table~\ref{tab:ablation_expert}, each specialized module progressively enhances the model's reasoning accuracy by addressing specific modality challenges.

\begin{table}[ht]
\centering
\small
\caption{Ablation of the three Expert modules. Each expert addresses specific challenges: Text/Image Experts mitigate factual hallucinations via RAG, while the Video Expert enhances temporal reasoning.}
\label{tab:ablation_expert}
\begin{tabular}{@{}lc@{}}
\toprule
Method Component & Accuracy (\%) \\ \midrule
VFB & 0.78 \\
\quad + Text Expert  & 0.83 \\
\quad + Image Expert  & 0.85 \\
\quad + Video Expert& \textbf{0.91} \\ \bottomrule
\end{tabular}
\end{table}

Text and Image Experts for Fact Verification. To mitigate modality interference and factual hallucinations prevalent in standard VLMs, the Text and Image Experts integrate structured metadata retrieval from SoccerWiki with visual similarity search. This retrieval-augmented mechanism grounds entity recognition (e.g., player identities and club history) in external knowledge, improving accuracy from \textbf{0.78 to 0.85}. These results demonstrate the efficacy of incorporating domain-specific knowledge for specialized sports analysis.

Video Expert for Semantic Alignment. The integration of the Video Expert further increases performance to \textbf{0.91}. Addressing the semantic ambiguity inherent in short SoccerNet queries, we employ a coarse-to-fine reasoning strategy. This pipeline first generates dense descriptions of player actions as intermediate representations, subsequently aligning low-level visual features with high-level semantic queries. Empirical results indicate that structured reasoning prior to answer generation significantly enhances performance in complex video VQA tasks.

{
    \small
    \bibliographystyle{ieeenat_fullname}
    \bibliography{main}

@misc{qwen25,
      title={Qwen2.5-VL Technical Report}, 
      author={Shuai Bai and Keqin Chen and Xuejing Liu and Jialin Wang and Wenbin Ge and Sibo Song and Kai Dang and Peng Wang and Shijie Wang and Jun Tang and Humen Zhong and Yuanzhi Zhu and Mingkun Yang and Zhaohai Li and Jianqiang Wan and Pengfei Wang and Wei Ding and Zheren Fu and Yiheng Xu and Jiabo Ye and Xi Zhang and Tianbao Xie and Zesen Cheng and Hang Zhang and Zhibo Yang and Haiyang Xu and Junyang Lin},
      year={2025},
      eprint={2502.13923},
      archivePrefix={arXiv},
      primaryClass={cs.CV},
      url={https://arxiv.org/abs/2502.13923}, 
}

@article{gemini,
  title={Gemini 2.5: Pushing the frontier with advanced reasoning, multimodality, long context, and next generation agentic capabilities},
  author={Comanici, Gheorghe and Bieber, Eric and Schaekermann, Mike and Pasupat, Ice and Sachdeva, Noveen and Dhillon, Inderjit and Blistein, Marcel and Ram, Ori and Zhang, Dan and Rosen, Evan and others},
  journal={arXiv preprint arXiv:2507.06261},
  year={2025}
}

@article{deepseekv32,
  title={Deepseek-v3. 2: Pushing the frontier of open large language models},
  author={Liu, Aixin and Mei, Aoxue and Lin, Bangcai and Xue, Bing and Wang, Bingxuan and Xu, Bingzheng and Wu, Bochao and Zhang, Bowei and Lin, Chaofan and Dong, Chen and others},
  journal={arXiv preprint arXiv:2512.02556},
  year={2025}
}

@misc{qwen3,
      title={Qwen3-VL Technical Report}, 
      author={Shuai Bai and Yuxuan Cai and Ruizhe Chen and Keqin Chen and Xionghui Chen and Zesen Cheng and Lianghao Deng and Wei Ding and Chang Gao and Chunjiang Ge and Wenbin Ge and Zhifang Guo and Qidong Huang and Jie Huang and Fei Huang and Binyuan Hui and Shutong Jiang and Zhaohai Li and Mingsheng Li and Mei Li and Kaixin Li and Zicheng Lin and Junyang Lin and Xuejing Liu and Jiawei Liu and Chenglong Liu and Yang Liu and Dayiheng Liu and Shixuan Liu and Dunjie Lu and Ruilin Luo and Chenxu Lv and Rui Men and Lingchen Meng and Xuancheng Ren and Xingzhang Ren and Sibo Song and Yuchong Sun and Jun Tang and Jianhong Tu and Jianqiang Wan and Peng Wang and Pengfei Wang and Qiuyue Wang and Yuxuan Wang and Tianbao Xie and Yiheng Xu and Haiyang Xu and Jin Xu and Zhibo Yang and Mingkun Yang and Jianxin Yang and An Yang and Bowen Yu and Fei Zhang and Hang Zhang and Xi Zhang and Bo Zheng and Humen Zhong and Jingren Zhou and Fan Zhou and Jing Zhou and Yuanzhi Zhu and Ke Zhu},
      year={2025},
      eprint={2511.21631},
      archivePrefix={arXiv},
      primaryClass={cs.CV},
      url={https://arxiv.org/abs/2511.21631}, 
}

@misc{internvl35,
      title={InternVL3.5: Advancing Open-Source Multimodal Models in Versatility, Reasoning, and Efficiency}, 
      author={Weiyun Wang and Zhangwei Gao and Lixin Gu and Hengjun Pu and Long Cui and Xingguang Wei and Zhaoyang Liu and Linglin Jing and Shenglong Ye and Jie Shao and Zhaokai Wang and Zhe Chen and Hongjie Zhang and Ganlin Yang and Haomin Wang and Qi Wei and Jinhui Yin and Wenhao Li and Erfei Cui and Guanzhou Chen and Zichen Ding and Changyao Tian and Zhenyu Wu and Jingjing Xie and Zehao Li and Bowen Yang and Yuchen Duan and Xuehui Wang and Zhi Hou and Haoran Hao and Tianyi Zhang and Songze Li and Xiangyu Zhao and Haodong Duan and Nianchen Deng and Bin Fu and Yinan He and Yi Wang and Conghui He and Botian Shi and Junjun He and Yingtong Xiong and Han Lv and Lijun Wu and Wenqi Shao and Kaipeng Zhang and Huipeng Deng and Biqing Qi and Jiaye Ge and Qipeng Guo and Wenwei Zhang and Songyang Zhang and Maosong Cao and Junyao Lin and Kexian Tang and Jianfei Gao and Haian Huang and Yuzhe Gu and Chengqi Lyu and Huanze Tang and Rui Wang and Haijun Lv and Wanli Ouyang and Limin Wang and Min Dou and Xizhou Zhu and Tong Lu and Dahua Lin and Jifeng Dai and Weijie Su and Bowen Zhou and Kai Chen and Yu Qiao and Wenhai Wang and Gen Luo},
      year={2025},
      eprint={2508.18265},
      archivePrefix={arXiv},
      primaryClass={cs.CV},
      url={https://arxiv.org/abs/2508.18265}, 
}

@misc{step3vl,
      title={STEP3-VL-10B Technical Report}, 
      author={Ailin Huang and Chengyuan Yao and Chunrui Han and Fanqi Wan and Hangyu Guo and Haoran Lv and Hongyu Zhou and Jia Wang and Jian Zhou and Jianjian Sun and Jingcheng Hu and Kangheng Lin and Liang Zhao and Mitt Huang and Song Yuan and Wenwen Qu and Xiangfeng Wang and Yanlin Lai and Yingxiu Zhao and Yinmin Zhang and Yukang Shi and Yuyang Chen and Zejia Weng and Ziyang Meng and Ang Li and Aobo Kong and Bo Dong and Changyi Wan and David Wang and Di Qi and Dingming Li and En Yu and Guopeng Li and Haiquan Yin and Han Zhou and Hanshan Zhang and Haolong Yan and Hebin Zhou and Hongbo Peng and Jiaran Zhang and Jiashu Lv and Jiayi Fu and Jie Cheng and Jie Zhou and Jisheng Yin and Jingjing Xie and Jingwei Wu and Jun Zhang and Junfeng Liu and Kaijun Tan and Kaiwen Yan and Liangyu Chen and Lina Chen and Mingliang Li and Qian Zhao and Quan Sun and Shaoliang Pang and Shengjie Fan and Shijie Shang and Siyuan Zhang and Tianhao You and Wei Ji and Wuxun Xie and Xiaobo Yang and Xiaojie Hou and Xiaoran Jiao and Xiaoxiao Ren and Xiangwen Kong and Xin Huang and Xin Wu and Xing Chen and Xinran Wang and Xuelin Zhang and Yana Wei and Yang Li and Yanming Xu and Yeqing Shen and Yuang Peng and Yue Peng and Yu Zhou and Yusheng Li and Yuxiang Yang and Yuyang Zhang and Zhe Xie and Zhewei Huang and Zhenyi Lu and Zhimin Fan and Zihui Cheng and Daxin Jiang and Qi Han and Xiangyu Zhang and Yibo Zhu and Zheng Ge},
      year={2026},
      eprint={2601.09668},
      archivePrefix={arXiv},
      primaryClass={cs.CV},
      url={https://arxiv.org/abs/2601.09668}, 
}

@article{lora,
  title={Lora: Low-rank adaptation of large language models.},
  author={Hu, Edward J and Shen, Yelong and Wallis, Phillip and Allen-Zhu, Zeyuan and Li, Yuanzhi and Wang, Shean and Wang, Liang and Chen, Weizhu and others},
  journal={Iclr},
  volume={1},
  number={2},
  pages={3},
  year={2022}
}

@misc{openai2024gpt4ocard,
      title={GPT-4o System Card}, 
      author={OpenAI and : and Aaron Hurst and Adam Lerer and Adam P. Goucher and Adam Perelman and Aditya Ramesh and Aidan Clark and AJ Ostrow and Akila Welihinda and Alan Hayes and Alec Radford and Aleksander Mądry and Alex Baker-Whitcomb and Alex Beutel and Alex Borzunov and Alex Carney and Alex Chow and Alex Kirillov and Alex Nichol and Alex Paino and Alex Renzin and Alex Tachard Passos and Alexander Kirillov and Alexi Christakis and Alexis Conneau and Ali Kamali and Allan Jabri and Allison Moyer and Allison Tam and Amadou Crookes and Amin Tootoochian and Amin Tootoonchian and Ananya Kumar and Andrea Vallone and Andrej Karpathy and Andrew Braunstein and Andrew Cann and Andrew Codispoti and Andrew Galu and Andrew Kondrich and Andrew Tulloch and Andrey Mishchenko and Angela Baek and Angela Jiang and Antoine Pelisse and Antonia Woodford and Anuj Gosalia and Arka Dhar and Ashley Pantuliano and Avi Nayak and Avital Oliver and Barret Zoph and Behrooz Ghorbani and Ben Leimberger and Ben Rossen and Ben Sokolowsky and Ben Wang and Benjamin Zweig and Beth Hoover and Blake Samic and Bob McGrew and Bobby Spero and Bogo Giertler and Bowen Cheng and Brad Lightcap and Brandon Walkin and Brendan Quinn and Brian Guarraci and Brian Hsu and Bright Kellogg and Brydon Eastman and Camillo Lugaresi and Carroll Wainwright and Cary Bassin and Cary Hudson and Casey Chu and Chad Nelson and Chak Li and Chan Jun Shern and Channing Conger and Charlotte Barette and Chelsea Voss and Chen Ding and Cheng Lu and Chong Zhang and Chris Beaumont and Chris Hallacy and Chris Koch and Christian Gibson and Christina Kim and Christine Choi and Christine McLeavey and Christopher Hesse and Claudia Fischer and Clemens Winter and Coley Czarnecki and Colin Jarvis and Colin Wei and Constantin Koumouzelis and Dane Sherburn and Daniel Kappler and Daniel Levin and Daniel Levy and David Carr and David Farhi and David Mely and David Robinson and David Sasaki and Denny Jin and Dev Valladares and Dimitris Tsipras and Doug Li and Duc Phong Nguyen and Duncan Findlay and Edede Oiwoh and Edmund Wong and Ehsan Asdar and Elizabeth Proehl and Elizabeth Yang and Eric Antonow and Eric Kramer and Eric Peterson and Eric Sigler and Eric Wallace and Eugene Brevdo and Evan Mays and Farzad Khorasani and Felipe Petroski Such and Filippo Raso and Francis Zhang and Fred von Lohmann and Freddie Sulit and Gabriel Goh and Gene Oden and Geoff Salmon and Giulio Starace and Greg Brockman and Hadi Salman and Haiming Bao and Haitang Hu and Hannah Wong and Haoyu Wang and Heather Schmidt and Heather Whitney and Heewoo Jun and Hendrik Kirchner and Henrique Ponde de Oliveira Pinto and Hongyu Ren and Huiwen Chang and Hyung Won Chung and Ian Kivlichan and Ian O'Connell and Ian O'Connell and Ian Osband and Ian Silber and Ian Sohl and Ibrahim Okuyucu and Ikai Lan and Ilya Kostrikov and Ilya Sutskever and Ingmar Kanitscheider and Ishaan Gulrajani and Jacob Coxon and Jacob Menick and Jakub Pachocki and James Aung and James Betker and James Crooks and James Lennon and Jamie Kiros and Jan Leike and Jane Park and Jason Kwon and Jason Phang and Jason Teplitz and Jason Wei and Jason Wolfe and Jay Chen and Jeff Harris and Jenia Varavva and Jessica Gan Lee and Jessica Shieh and Ji Lin and Jiahui Yu and Jiayi Weng and Jie Tang and Jieqi Yu and Joanne Jang and Joaquin Quinonero Candela and Joe Beutler and Joe Landers and Joel Parish and Johannes Heidecke and John Schulman and Jonathan Lachman and Jonathan McKay and Jonathan Uesato and Jonathan Ward and Jong Wook Kim and Joost Huizinga and Jordan Sitkin and Jos Kraaijeveld and Josh Gross and Josh Kaplan and Josh Snyder and Joshua Achiam and Joy Jiao and Joyce Lee and Juntang Zhuang and Justyn Harriman and Kai Fricke and Kai Hayashi and Karan Singhal and Katy Shi and Kavin Karthik and Kayla Wood and Kendra Rimbach and Kenny Hsu and Kenny Nguyen and Keren Gu-Lemberg and Kevin Button and Kevin Liu and Kiel Howe and Krithika Muthukumar and Kyle Luther and Lama Ahmad and Larry Kai and Lauren Itow and Lauren Workman and Leher Pathak and Leo Chen and Li Jing and Lia Guy and Liam Fedus and Liang Zhou and Lien Mamitsuka and Lilian Weng and Lindsay McCallum and Lindsey Held and Long Ouyang and Louis Feuvrier and Lu Zhang and Lukas Kondraciuk and Lukasz Kaiser and Luke Hewitt and Luke Metz and Lyric Doshi and Mada Aflak and Maddie Simens and Madelaine Boyd and Madeleine Thompson and Marat Dukhan and Mark Chen and Mark Gray and Mark Hudnall and Marvin Zhang and Marwan Aljubeh and Mateusz Litwin and Matthew Zeng and Max Johnson and Maya Shetty and Mayank Gupta and Meghan Shah and Mehmet Yatbaz and Meng Jia Yang and Mengchao Zhong and Mia Glaese and Mianna Chen and Michael Janner and Michael Lampe and Michael Petrov and Michael Wu and Michele Wang and Michelle Fradin and Michelle Pokrass and Miguel Castro and Miguel Oom Temudo de Castro and Mikhail Pavlov and Miles Brundage and Miles Wang and Minal Khan and Mira Murati and Mo Bavarian and Molly Lin and Murat Yesildal and Nacho Soto and Natalia Gimelshein and Natalie Cone and Natalie Staudacher and Natalie Summers and Natan LaFontaine and Neil Chowdhury and Nick Ryder and Nick Stathas and Nick Turley and Nik Tezak and Niko Felix and Nithanth Kudige and Nitish Keskar and Noah Deutsch and Noel Bundick and Nora Puckett and Ofir Nachum and Ola Okelola and Oleg Boiko and Oleg Murk and Oliver Jaffe and Olivia Watkins and Olivier Godement and Owen Campbell-Moore and Patrick Chao and Paul McMillan and Pavel Belov and Peng Su and Peter Bak and Peter Bakkum and Peter Deng and Peter Dolan and Peter Hoeschele and Peter Welinder and Phil Tillet and Philip Pronin and Philippe Tillet and Prafulla Dhariwal and Qiming Yuan and Rachel Dias and Rachel Lim and Rahul Arora and Rajan Troll and Randall Lin and Rapha Gontijo Lopes and Raul Puri and Reah Miyara and Reimar Leike and Renaud Gaubert and Reza Zamani and Ricky Wang and Rob Donnelly and Rob Honsby and Rocky Smith and Rohan Sahai and Rohit Ramchandani and Romain Huet and Rory Carmichael and Rowan Zellers and Roy Chen and Ruby Chen and Ruslan Nigmatullin and Ryan Cheu and Saachi Jain and Sam Altman and Sam Schoenholz and Sam Toizer and Samuel Miserendino and Sandhini Agarwal and Sara Culver and Scott Ethersmith and Scott Gray and Sean Grove and Sean Metzger and Shamez Hermani and Shantanu Jain and Shengjia Zhao and Sherwin Wu and Shino Jomoto and Shirong Wu and Shuaiqi and Xia and Sonia Phene and Spencer Papay and Srinivas Narayanan and Steve Coffey and Steve Lee and Stewart Hall and Suchir Balaji and Tal Broda and Tal Stramer and Tao Xu and Tarun Gogineni and Taya Christianson and Ted Sanders and Tejal Patwardhan and Thomas Cunninghman and Thomas Degry and Thomas Dimson and Thomas Raoux and Thomas Shadwell and Tianhao Zheng and Todd Underwood and Todor Markov and Toki Sherbakov and Tom Rubin and Tom Stasi and Tomer Kaftan and Tristan Heywood and Troy Peterson and Tyce Walters and Tyna Eloundou and Valerie Qi and Veit Moeller and Vinnie Monaco and Vishal Kuo and Vlad Fomenko and Wayne Chang and Weiyi Zheng and Wenda Zhou and Wesam Manassra and Will Sheu and Wojciech Zaremba and Yash Patil and Yilei Qian and Yongjik Kim and Youlong Cheng and Yu Zhang and Yuchen He and Yuchen Zhang and Yujia Jin and Yunxing Dai and Yury Malkov},
      year={2024},
      eprint={2410.21276},
      archivePrefix={arXiv},
      primaryClass={cs.CL},
      url={https://arxiv.org/abs/2410.21276}, 
}

@inproceedings{soccerbench,
      title = {Multi-Agent System for Comprehensive Soccer Understanding},
      author = {Rao, Jiayuan and Li, Zifeng and Wu, Haoning and Zhang, Ya and Wang, Yanfeng and Xie, Weidi},
      booktitle = {ACM Multimedia 2025},
      year = {2025}
}

@inproceedings{soccernet,
   title={SoccerNet: A Scalable Dataset for Action Spotting in Soccer Videos},
   url={http://dx.doi.org/10.1109/CVPRW.2018.00223},
   DOI={10.1109/cvprw.2018.00223},
   booktitle={2018 IEEE/CVF Conference on Computer Vision and Pattern Recognition Workshops (CVPRW)},
   publisher={IEEE},
   author={Giancola, Silvio and Amine, Mohieddine and Dghaily, Tarek and Ghanem, Bernard},
   year={2018},
   month=jun }

@misc{soccernetv2,
      title={SoccerNet-v2: A Dataset and Benchmarks for Holistic Understanding of Broadcast Soccer Videos}, 
      author={Adrien Deliège and Anthony Cioppa and Silvio Giancola and Meisam J. Seikavandi and Jacob V. Dueholm and Kamal Nasrollahi and Bernard Ghanem and Thomas B. Moeslund and Marc Van Droogenbroeck},
      year={2021},
      eprint={2011.13367},
      archivePrefix={arXiv},
      primaryClass={cs.CV}
}

@misc{soccernettracking,
      title={SoccerNet-Tracking: Multiple Object Tracking Dataset and Benchmark in Soccer Videos}, 
      author={Anthony Cioppa and Silvio Giancola and Adrien Deliege and Le Kang and Xin Zhou and Zhiyu Cheng and Bernard Ghanem and Marc Van Droogenbroeck},
      year={2022},
      eprint={2204.06918},
      archivePrefix={arXiv},
      primaryClass={cs.CV}
}

@article{Cioppa2022,
  title={Scaling up SoccerNet with multi-view spatial localization and re-identification},
  author={Cioppa, Anthony and Deli{\`e}ge, Adrien and Giancola, Silvio and Ghanem, Bernard and Van Droogenbroeck, Marc},
  journal={Scientific Data},
  year={2022},
  volume={9},
  number={1},
  pages={355},
}

@article{soccerchat,
  author = {Gautam, Sushant and Midoglu, Cise and Thambawita, Vajira and Riegler, Michael A. and Halvorsen, Pål and Shah, Mubarak},
  title = {SoccerChat: Integrating Multimodal Data for Enhanced Soccer Game Understanding},
  journal = {ArXiv e-prints},
  year = {2025},
  month = may,
  eprint = {2505.16630},
  doi = {10.48550/arXiv.2505.16630}
}

@inproceedings{soccerreplay,
    title   = {Towards Universal Soccer Video Understanding},
    author  = {Rao, Jiayuan and Wu, Haoning and Jiang, Hao and Zhang, Ya and Wang, Yanfeng and Xie, Weidi},
    booktitle = {Proceedings of the IEEE/CVF Conference on Computer Vision and Pattern Recognition (CVPR)},
    year    = {2025},
}

@inproceedings{sn-caption,
  title={SoccerNet-caption: Dense video captioning for soccer broadcasts commentaries},
  author={Mkhallati, Hassan and Cioppa, Anthony and Giancola, Silvio and Ghanem, Bernard and Van Droogenbroeck, Marc},
  booktitle={Proceedings of the IEEE/CVF Conference on Computer Vision and Pattern Recognition},
  pages={5074--5085},
  year={2023}
}
}


\end{document}